\documentclass[runningheads]{llncs}
\usepackage{graphicx}

\usepackage{tikz}
\usepackage{comment}
\usepackage{amsmath,amssymb} 
\usepackage{color}
\usepackage{subfigure}

\usepackage[accsupp]{axessibility}  

\begin{document}
\pagestyle{headings}
\mainmatter
\def\ECCVSubNumber{5158}  

\title{Camera Pose Auto-Encoders for Improving Pose Regression} 

\titlerunning{Camera Pose Auto-Encoders for Improving Pose Regression}
%
\author{Yoli Shavit \and
Yosi Keller}
\authorrunning{Y. Shavit and Y. Keller}
%
\institute{Bar-Ilan University, Ramat Gan, Israel\\
\email{\{yolisha, yosi.keller\}@gmail.com}}

\maketitle

\begin{abstract}
Absolute pose regressor (APR) networks are trained to estimate the pose of
the camera given a captured image. They compute latent image
representations from which the camera position and orientation are
regressed. APRs provide a different tradeoff between localization accuracy,
runtime, and memory, compared to structure-based localization schemes that
provide state-of-the-art accuracy. In this work, we introduce Camera Pose
Auto-Encoders (PAEs), multilayer perceptrons that are trained via a
Teacher-Student approach to encode camera poses using APRs as their
teachers. We show that the resulting latent pose representations can closely
reproduce APR performance and demonstrate their effectiveness for related
tasks. Specifically, we propose a light-weight test-time optimization in
which the closest train poses are encoded and used to refine camera position
estimation. This procedure achieves a new state-of-the-art position accuracy
for APRs, on both the CambridgeLandmarks and 7Scenes benchmarks. We also
show that train images can be reconstructed from the learned pose encoding,
paving the way for integrating visual information from the train set at a
low memory cost. Our code and pre-trained models are available at \url{https://github.com/yolish/camera-pose-auto-encoders}.
\end{abstract}

\section{Introduction}

Estimating the position and orientation of a camera given a query image is a
fundamental problem in computer vision. It has applications in multiple
domains, such as virtual and augmented reality, indoor navigation,
autonomous driving, to name a few. Contemporary state-of-the-art camera
localization methods are based on matching pixels in the query image to 3D world
coordinates. Such 2D-3D correspondences are obtained either through scene
coordinate regression \cite{DSAC,DSAC++,9394752} or by extracting and matching deep
features in the query and reference images, for which 3D information is
available \cite{taira2018inloc,sarlin2019coarse,noh2017large,dusmanu2019d2}.
The resulting correspondences are used to estimate the camera pose with
Perspective-N-Point (PnP) and RANSAC \cite{fischler1981random}.
Consequently, both approaches require the intrinsic parameters of the query
camera, which might not be available or accurate. In addition, matching the
query and reference images typically involves storing visual and 3D
information on a remote server or the end device.

An alternative approach is to directly regress the camera pose from the
query image \cite{kendall2015posenet} with absolute pose regressors (APRs).
With these methods, the query image is first encoded into a latent
representation using a convolutional backbone \cite%
{melekhov2017image,naseer2017deep,walch2017image,wu2017delving,shavitferensirpnet,wang2020atloc,cai2019hybrid}
or Transformers encoders \cite{ShavitFerensIccv21}. The latent image
representation is then used to regress the position and orientation with
one or more multi-layer perceptron heads. APRs are typically optimized
through a supervision of the ground truth poses \cite%
{kendall2015posenet,shavit2019introduction,kendall2017geometric} and can be
trained per scene, or as more recently proposed, in a multi-scene manner
(training a single model for multiple scenes) \cite%
{ShavitFerensIccv21,blanton2020extending}. While being less accurate than
state-of-the-art (SOTA) structure-based localization approaches \cite%
{DSAC,DSAC++}, APRs offer a different trade-off between accuracy versus
runtime and memory, by being faster and simpler. In addition, they do not
require the intrinsic parameters of the query camera as an input. A related body of work
focuses on regressing the relative motion between a pair of images. When the
camera pose of a reference image is known, its relative motion to the query
can be used to estimate its pose by simple matrix inversion and
multiplication. By harnessing relative pose regression for camera pose
estimation, relative pose regressors (RPRs) can offer better generalization
and accuracy \cite{ding2019camnet} but require images or their
model-specific high-dimensional encoding to be available at inference
time (supplementary section 1.1). Although RPRs can also be coupled with a sequential acquisition, we are
mainly interested in scenarios where only a single query image is provided
at a time.
\begin{figure}[tbh]
\centering
\vspace{-1.5em} \includegraphics[scale=0.45]{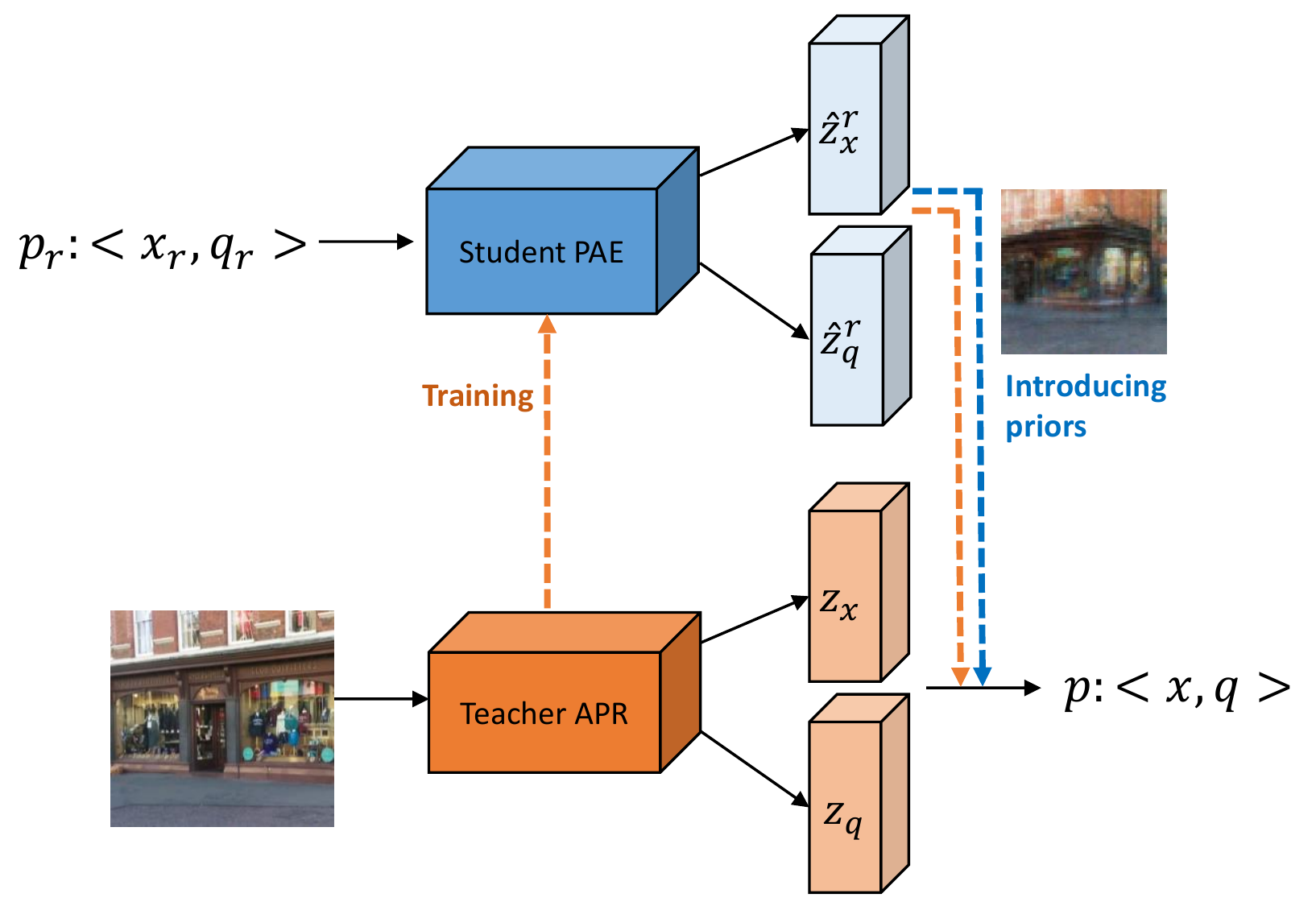}
\caption{A Camera Pose Auto-Encoder (PAE) is trained using a Teacher-Student
approach, to generate the same  pose encoding as the one computed by a teacher APR, enabling
the teacher to perform accurate pose regression. The trained student PAE allows to introduce prior information and improve the
teacher APR localization accuracy.} \vspace{-2em}
\label{fig:teaser}
\end{figure}

In this work, we propose to make the geometric and visual information of
reference images (training set) available during inference time, without
incurring significant memory or runtime costs. Our motivation is to maintain
the attractive properties of APRs (fast, lightweight, and standalone) while
improving their localization accuracy using prior information. For this
purpose, we propose the Camera Pose Auto-Encoder (PAE) shown in Fig. \ref%
{fig:teaser}: an MLP that is trained to encode camera poses into latent
representations learned by APRs from the respective images. We train PAEs using a Teacher-Student approach, where given a latent representation of an
image, obtained with a pretrained teacher APR, the student PAE learns to
generate the same encoding for the respective camera pose. The pose encoding
is optimized to be as similar as possible to the latent image representation
and to enable accurate pose regression with the teacher APR. The proposed training scheme uses multiple images acquired from similar poses with varying
appearances, but the PAE is applied without using the reference image as input.
Thus, the resulting PAE-based pose encoding is robust to appearance. Once a
PAE is trained, we can use it to introduce prior information and improve the
APR localization accuracy.

We evaluate our approach on the Cambridge Landmarks and 7Scenes datasets,
which provide various outdoor and indoor localization challenges. We first
show that student PAEs can closely reproduce the performance of their
teacher APRs across datasets, APR and PAE architectures. We then provide examples for using PAEs to improve camera pose regression.
We describe a lightweight test-time optimization method,
where given an initial pose estimate, the nearest poses in the train set
can be encoded and used to derive an improved \textit{position} estimation. This
simple procedure achieves a new state-of-the-art localization accuracy
compared to current APR solutions across datasets. We further show that
images can be reconstructed from camera pose encoding, allowing for
performing relative pose regression without the need to store the actual
images or their model-specific encodings. This in turn results in competitive
position estimation and improves the initial estimate of the teacher APR.

In summary, our main contributions are as follows:

\begin{itemize}
\item We introduce a Teacher-Student approach for learning to encode poses
into appearance-robust informative latent representations, and show that the
trained student Camera Pose Auto Encoders (PAEs) effectively reproduce their
teacher APRs.

\item We propose a fast and lightweight test-time optimization procedure
which utilizes PAEs and achieves a new state-of-the-art position accuracy
for absolute pose regression.

\item We show that the learned camera pose encoding can be used for image
reconstruction, paving the way for coupling relative and absolute pose regression and improving pose
estimation, without the typical memory burden of RPRs.
\end{itemize}

\section{Related Work}

\label{sec:related}

\subsection{Structure-based Pose Estimation}

\label{subsec:related-Features-Points} Structure-based pose estimation
methods detect or estimate either 2D or 3D feature points that are matched
to a set of reference 3D coordinates. PnP approaches are then applied to
estimate the camera pose based on 2D-to-3D matches \cite{taira2018inloc}%
. The 3D\ scene model is commonly acquired using SfM \cite%
{sattler2016efficient}, or a depth sensor \cite%
{DBLP:conf/cvpr/CavallariGLVST17}. Such approaches achieve SOTA localization
accuracy but require the ground truth poses and 3D coordinates of a set of
reference images and their respective local features, as well as the
intrinsics parameters of the query and reference cameras. They also need to
store the image descriptors for retrieving the reference images that will be
matched and the 3D coordinates of their local features. The required memory
can be reduced by product quantization of the 3D point descriptors \cite%
{Torii}, or using only a subset of all 3D points \cite%
{Trujillo,sattler2016efficient}. This subset can be obtained, for example,
by a prioritized matching step that first considers features more likely to
yield valid 2D-to-3D matches \cite{sattler2016efficient}. Recently, Sarlin
et al. \cite{sarlin21pixloc} proposed a CNN to detect multilevel invariant
visual features, with pixel-wise confidence for query and reference images.
Levenberg-Marquardt optimization was applied in a coarse-to-fine manner, to match the
corresponding features using their confidence, and the training was
supervised by the predicted pose. Instead of retrieving reference images and
matching local features to obtain 2D-to-3D correspondences, some approaches
regress the 3D scene coordinates directly from the query image \cite%
{shotton2013scene}. The resulting matches between 2D pixels and 3D
coordinates regressed from the query image are used to estimate the pose
with PnP-RANSAC. Brachmann and Rother \cite{DSAC,DSAC++} extended this
approach by training an end-to-end trainable network. A CNN was used to
estimate the 3D locations corresponding to the pixels in the query image,
and the 2D-to-3D correspondences were used by a differentiable
PnP-RANSAC to estimate the camera pose. Such approaches achieve
state-of-the-art accuracy, but similarly to other structure-based pose
estimation methods, require the intrinsics of the query camera. \vspace{-1em}

\subsection{Regression-based Pose Estimation}

\label{subsec:related-regression-based}

Kendall et al. \cite{kendall2015posenet} were the first to apply
convolutional backbones to absolute pose regression, where the camera
pose is directly regressed from the query image. Specifically, an MLP head
was attached to a GoogLeNet backbone, to regress the camera's position and
orientation. Regression-based approaches are far less accurate than SOTA
structure-based localization \cite{DSAC,DSAC++}, but allow pose estimation
with a single forward pass in just a few milliseconds and without requiring
the camera intrinsics, which might be inaccurate and unavailable. Some APR formulations proposed using different CNN backbones \cite%
{melekhov2017image,naseer2017deep,wu2017delving,shavitferensirpnet} and
deeper architectures for the MLP head \cite{wu2017delving,naseer2017deep}.
Other works tried to reduce overfitting by averaging predictions from
models with randomly dropped activations \cite{kendall2016modelling} or by
reducing the dimensionality of the global image encoding with
Long-Short-Term-Memory (LSTM) layers \cite{walch2017image}.
Multimodality fusion (for example, with inertial sensors) was also
suggested as a means of improving accuracy \cite{brahmbhatt2018geometry}.
Attention-based schemes and Transformers were more recently shown to boost
the performance of APRs. Wang et al. suggested to use attention to guide the
regression process \cite{wang2020atloc}. Dot product self-attention was
applied to the output of the CNN backbone and updated with the new
representation based on attention (by summation). The pose was then
regressed with an MLP head. A transformer-based approach to multiscene
absolute pose regression was proposed by Shavit et al. \cite%
{ShavitFerensIccv21}. In their work, the authors used a shared backbone to
encode multiple scenes using a full transformer. The scheme was shown to
provide SOTA multi-scene pose accuracy compared to current APRs.
One of the main challenges in APR is weighing the position and orientation
losses. Kendall et al. \cite{kendall2017geometric} learn the trade-off
between the losses to improve the localization accuracy. Although this approach
was adopted by many pose regressors, it requires manually tuning the
parameters' initialization for different datasets \cite{valada2018deep}. To
reduce the need for additional parameters while maintaining comparable
accuracy, Shavit et al. \cite{shavitferensirpnet}, trained separate models
for position and orientation. Other orientation formulations were proposed
to improve the pose loss balance and stability \cite%
{wu2017delving,brahmbhatt2018geometry}.

The relative motion between the query image and a reference image, for which
the ground truth pose is known, has also been employed to estimate the
absolute camera pose in a similar, yet separate subclass of works. Thus, learning such RPR models focuses on regressing the relative pose
given a pair of images \cite{balntas2018relocnet,ding2019camnet}. These
methods generalize better since the model is not restricted to an absolute
reference scene, but require a pose-labeled database of anchors at
inference time. Combining relative and absolute regression has been shown to
achieve impressive accuracy \cite{radwan2018vlocnet++,ding2019camnet}, but
requires the encoding of the train images or localization with more than a
single query image. As graph neural networks (GNNs) allow exchanging
information between non-consecutive frames of a video clip, researchers were
motivated to use them for learning multi-image RPR for absolute pose
estimation. Xue et al. \cite{9156582} introduced the GL-Net GNN for
multiframe learning, where an estimate of the relative pose loss is applied to
regularize the APR. Turkoglu et al. \cite{9665967} also applied GNN to multi-frame relative localization. In both the training and testing phases,
NetVLAD embeddings are used to retrieve the most similar images. A GNN is
applied to the retrieved images, and message passing is used to estimate the
pose of the camera. Visual landmarks were used by Saha et al. in the AnchorPoint
localization approach \cite{DBLP:conf/bmvc/SahaVJ18}. With this method,
anchor points are distributed uniformly throughout the environment to allow
the network to predict, when presented with a query image, which anchor
points will be the most relevant in addition to where they are located in
relation to the query image.

The inversion of the neural radiation field (NeRF) was
recently proposed for test-time optimization of camera poses \cite%
{yen2020inerf}. In the proposed scheme, the appearance deviation between the
input query and the rendered image was used to optimize the camera pose,
without requiring an explicit 3D scene representation (as NeRFs can be
estimated directly from images). While offering a novel and innovative
approach to camera pose estimation, this procedure is relatively slow
compared to structure and regression-based localization methods. In this
work, we focus on absolute pose regression with a single image. We aim at
maintaining the low memory and runtime requirements, while improving
accuracy through encoding of pose priors. \vspace{-1em}

\section{Absolute Pose Regression using Pose Auto-Encoders}

A camera pose $\mathbf{p}$, can be represented with the tuple $<\mathbf{x},%
\mathbf{q}>$ where $\mathbf{x}\in \mathbb{R}^{3}$ is the position of the
camera in world coordinates and $\mathbf{q}\in \mathbb{S}^{3}$ is a unit
quaternion encoding its spatial orientation. An APR $\mathbf{A}$ \cite%
{kendall2015posenet,shavit2019introduction,kendall2017geometric} can be
decomposed into the encoders $\mathbf{E_{x}}$ and $\mathbf{E_{q}}$, which encode
the\textit{\ query image} into respective latent representations $\mathbf{%
z_{x}}\in \mathbb{R}^{d}$ and $\mathbf{z_{q}}\in \mathbb{R}^{d}$, and the heads $%
\mathbf{R_{x}}$ and $\mathbf{R_{q}}$, which regress $\mathbf{x}$ and $%
\mathbf{q}$ from $\mathbf{z_{x}}$ and $\mathbf{z_{q}}$, respectively. In
this work we propose the \textit{camera pose auto-encoder} (PAE) $\mathbf{f}$%
, which encodes \textit{the pose} $<\mathbf{x},\mathbf{q}>$ to the
high-dimensional latent encodings, $\mathbf{\hat{z}_{x}}\in \mathbb{R}^{d}$
and $\mathbf{\hat{z}_{q}}\in \mathbb{R}^{d}$, respectively. We would like $%
\mathbf{\hat{z}_{x}}$ and $\mathbf{\hat{z}_{q}}$ to encode geometric and
visual information such that an APR's heads $\mathbf{R_{x}}$ and $\mathbf{%
R_{q}}$ can decode back $<\mathbf{x},\mathbf{q}>$. We show that PAE can
be applied to single- and multi-scene APRs.\vspace{-1em}

\subsection{Training Camera Pose Auto-Encoders}

\label{subsec:training} 
\begin{figure}[tbh]
\centering
\includegraphics[width=0.6\linewidth]{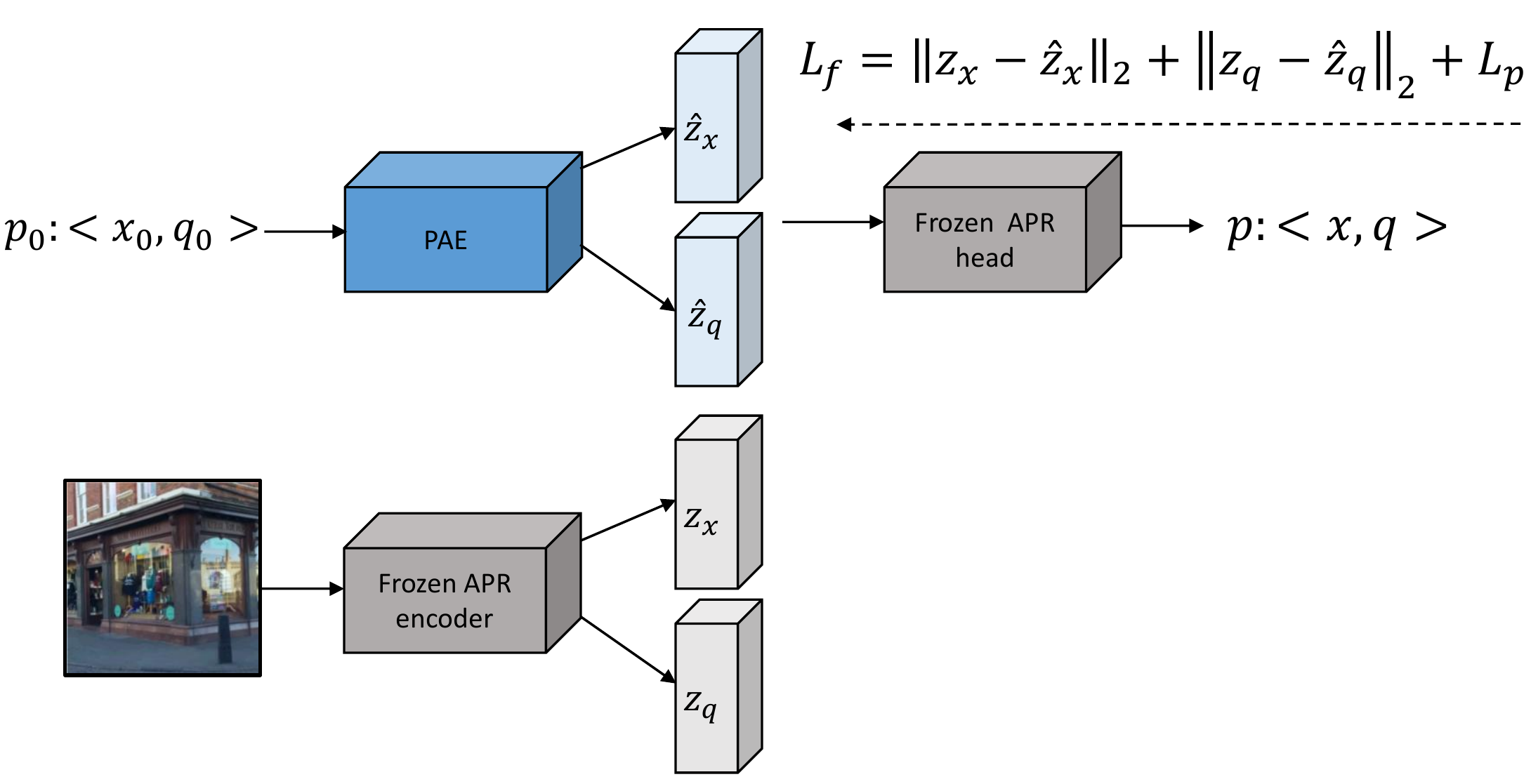}
\vspace{-1em}
\caption{A Teacher-Student approach for training PAEs. A trained APR teacher network is used to train the student PAE network.}
 \vspace{-2em}
\label{fig:training}
\end{figure}

An APR $\mathbf{A}$ plays a dual role in training $%
\mathbf{f}$, both as a teacher and as a decoder. Specifically, the PAE $%
\mathbf{f}$ can be considered as a student of $\mathbf{A,}$ such that $%
\mathbf{A}$'s outputs $\mathbf{z_{x}}$ and $\mathbf{z_{q}}$ are used to
train the PAE by minimizing the loss:
\begin{equation}
L_{\mathbf{f}}=||\mathbf{z_{x}}-\mathbf{\hat{z}_{x}}||_{2}+||\mathbf{z_{q}}-%
\mathbf{\hat{z}_{q}}||_{2}+L_{\mathbf{p}},  \label{equ:pose encoder loss}
\end{equation}%
where $\mathbf{\hat{z}_{x}}$ and $\mathbf{\hat{z}_{q}}$ are the outputs of the PAE. We require $\mathbf{\hat{z}_{x}}$ and $\mathbf{\hat{z}_{q}}$ to
allow an accurate \textit{decoding} of the pose $<\mathbf{x},\mathbf{q}>$
using the respective regressors $\mathbf{R_{x}}$ and $\mathbf{R_{q}}$, minimizing the loss of camera pose \cite{kendall2017geometric}, given by:
\begin{equation}
L_{\mathbf{p}}=L_{\mathbf{x}}\exp (-s_{\mathbf{x}})+s_{\mathbf{x}}+L_{%
\mathbf{q}}\exp (-s_{\mathbf{q}})+s_{\mathbf{q}}
\label{equ:learnable pose loss}
\end{equation}%
where $s_{x}$ and $s_{q}$ are learned parameters representing the uncertainty associated with position and orientation estimation, respectively, \cite{kendall2017geometric} and $L_{\mathbf{x}}$ and $L_{\mathbf{q}}$ are the position and orientation
losses, with respect to a ground truth pose $\mathbf{p}_{0}=<\mathbf{x}_{0},%
\mathbf{q}_{0}>$:%
\begin{equation}
L_{\mathbf{x}}=||\mathbf{x}_{0}-\mathbf{x}||_{2}  \label{equ:position loss}
\end{equation}%
and%
\begin{equation}
L_{\mathbf{q}}=||\mathbf{q_{0}}-\frac{\mathbf{q}}{||\mathbf{q}||}||_{2}.
\label{equ:orientation loss}
\end{equation}%
Following previous works \cite%
{kendall2015posenet,kendall2017geometric,ShavitFerensIccv21}, we normalize $%
\mathbf{q}$ to a unit norm quaternion to map it to a valid spatial rotation.
The training and formulation of $\mathbf{f}$ can be extended to multi-scene
APR by additionally encoding the scene index $\mathbf{s,}$ given as input.
Figure \ref{fig:training} illustrates the training process of PAEs.

\subsection{Network Architecture}

\label{subsec:PAE}

In this work, we implement a camera pose auto-encoder $\mathbf{f}$ using two
MLPs encoding $\mathbf{x}$ and $\mathbf{q}$, respectively. Following the
observations of \cite{rahaman2018spectral,tancik2020fourier} that high
frequency functions can help in learning low-dimensional signals (and in
particularly camera poses \cite{mildenhall2020nerf}), we first embed $%
\mathbf{x}$ and $\mathbf{q}$ in a higher dimensional space using Fourier
Features . We use the formulation and implementation of \cite%
{mildenhall2020nerf}, and apply the following function:
\begin{equation}
\gamma (p)=\left( \sin \left( 2^{0}\pi p\right) ,\cos \left( 2^{0}\pi
p\right) ,\cdots ,\sin \left( 2^{-1}\pi p\right) ,\cos \left( 2^{-1}\pi
p\right) \right),  \label{eq:fourier_features}
\end{equation}%
$\gamma $ maps $\mathbb{R}$ into a higher dimensional space $\mathbb{R}%
^{2L}, $ and is separately applied to each coordinate of $\textbf{x}$ and $\textbf{q}$,
respectively. We also concatenate the original input so that the final
dimension of the encoding is $2L+d_{0}$, $d_{0}$ being the dimension of
the embedded input. The corresponding MLP head is then applied on the
resulting representation to compute $\mathbf{e_{x}}\in \mathbb{R}^{d}$ and $%
\mathbf{e_{q}}\in \mathbb{R}^{d}$. In a multi-scene scenario with $n_{s}$
encoded scenes, a scene index $s=0,...,n_{s}-1$ is
encoded using Fourier Features as in Eq.~\ref{eq:fourier_features},
similarly to $\textbf{x}$ and $\textbf{q}$, and then concatenated to their encoding before
applying the respective MLP head. \vspace{-1em}

\subsection{Applications of Camera Pose Auto-Encoders}

\label{subsec:applications} PAEs allow us to introduce prior information
(i.e., localization parameters of the training set's poses) at a low memory and run-time cost to improve the
localization accuracy of APRs. We demonstrate this idea through two example
applications: Test-time Position Refinement and Virtual Relative Pose
Regression.

\subsubsection{Test-time Position Refinement}

\label{subsubsec:TT refine}
\begin{figure}[tbh]
\vspace{-2em} \centering
\includegraphics[scale=0.35]{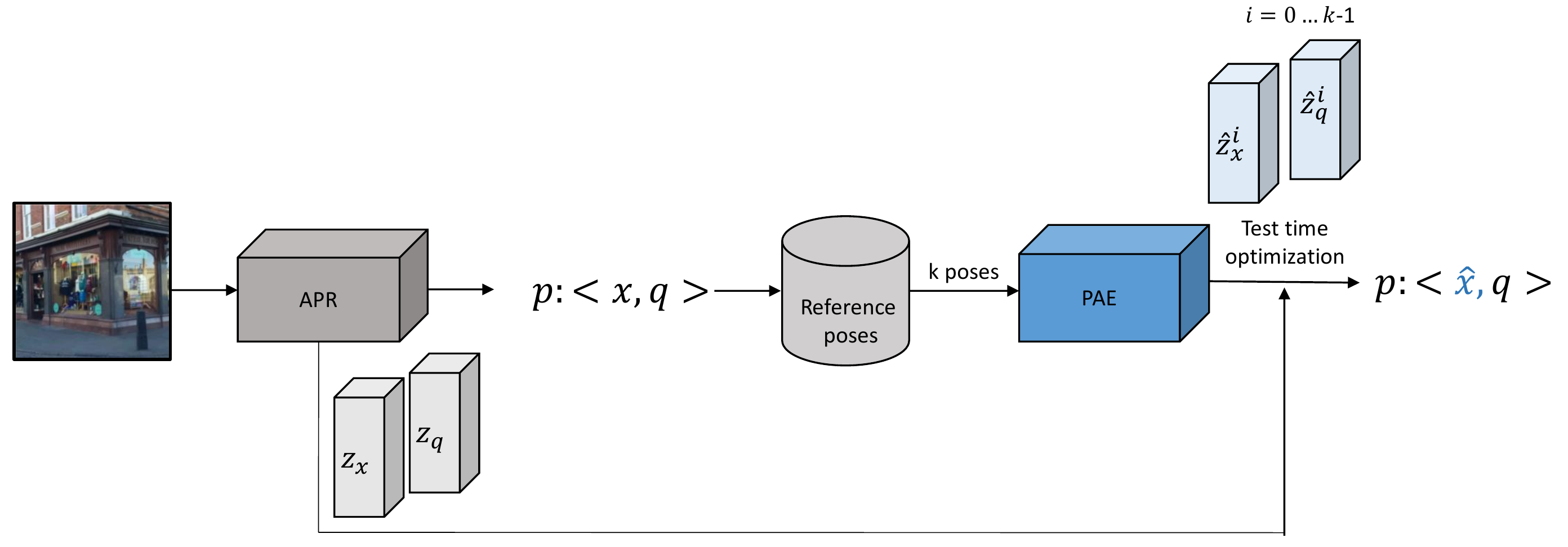} \vspace{-1em}
\caption{Test-time optimization of position estimation with PAEs.}
\vspace{-2em}
\label{fig:test-time-optimization}
\end{figure}

Given a pre-trained APR $\mathbf{A}$ and a query
image, we first compute the latent representations $\mathbf{z_{x}}$ and $%
\mathbf{z_{q}}$ and a pose estimate $\mathbf{p:<x,q>}$. Using $\mathbf{p}$,
we can get $k$ poses of images from the training set, whose \textit{%
poses} are the closest to the \textit{pose} of the query image. This
requires only to store the pose information $<\mathbf{x},\mathbf{q}>$ $\in
\mathbb{R}^{7}$, and not the images themselves. Given a pre-trained pose
auto-encoder $\mathbf{f}$, we encode each of the $k$ train reference poses, $%
\{p_{r}^{i}\}_{i=0}^{k-1}$, into latent representations: $\{\mathbf{\hat{z}%
_{x}^{i}},\mathbf{\hat{z}_{q}}^{i}\}_{i=0}^{k-1}$. Using the simple test-time
optimization shown in Fig. \ref{fig:test-time-optimization}, we can estimate
$\mathbf{x}$ as an affine combination of train positions:
\begin{equation}
\mathbf{x=}\sum_{i=0}^{k-1}a_{i}{x}_{r}^{i},\text{ }s.t.\sum a_{i}=1.
\end{equation}%
The weight vector $\mathbf{a}$ is calculated by optimizing an MLP
regressor for an affine combination of train pose encodings that are
closest to the latent encoding of the image
\begin{align*}
\mathbf{a}& =\arg \underset{\mathbf{a}}{\min }||\mathbf{z_{p}}%
-\sum_{i=0}^{k-1}a_{i}\mathbf{\hat{z}_{p_{r}}^{i}}||_{2}, \\
\text{ s.t.}\sum a_{i}& =1,\text{ }\mathbf{z_{p}=}%
\begin{bmatrix}
\mathbf{z_{x}} \\
\mathbf{z_{q}}%
\end{bmatrix}%
\end{align*}%
A similar test-time optimization was shown to perform well for estimating the
camera pose from the nearest image descriptors \cite%
{sattler2019understanding}. However, as opposed to poses, image descriptors
mostly encode the image appearance and are thus encoder dependent.

\subsubsection{Virtual Relative Pose Regression}

\label{subsubsec:virtual RPS} 
\begin{figure}[tbh]
\vspace{
-2em} \centering
\includegraphics[scale=0.35]{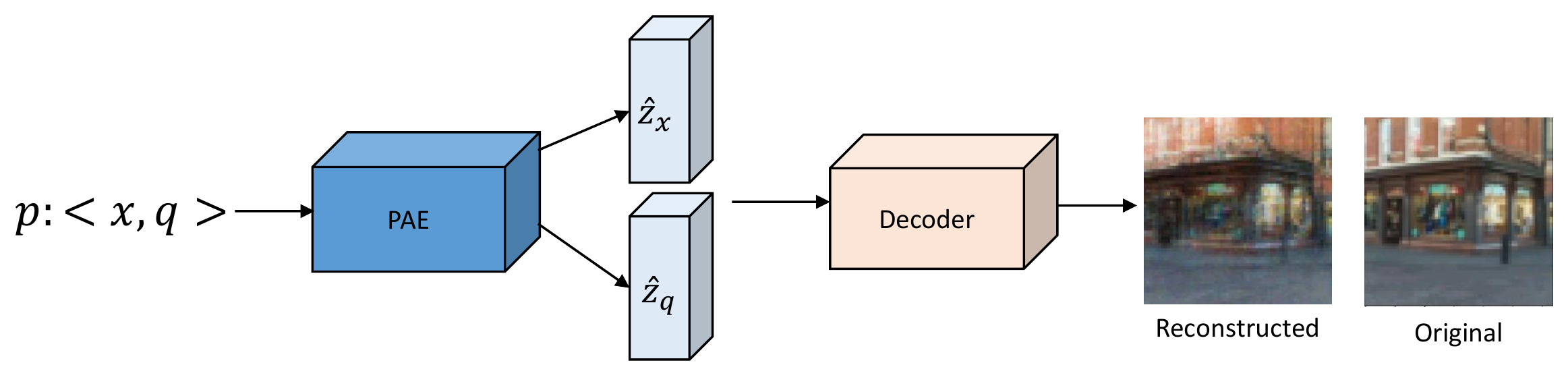} \vspace{-1em}
\caption{Decoding images from learned camera pose encoding.}
\vspace{-2em}
\label{fig:img-recon}
\end{figure}

\begin{figure}[tbh]
\centering
\includegraphics[scale=0.35]{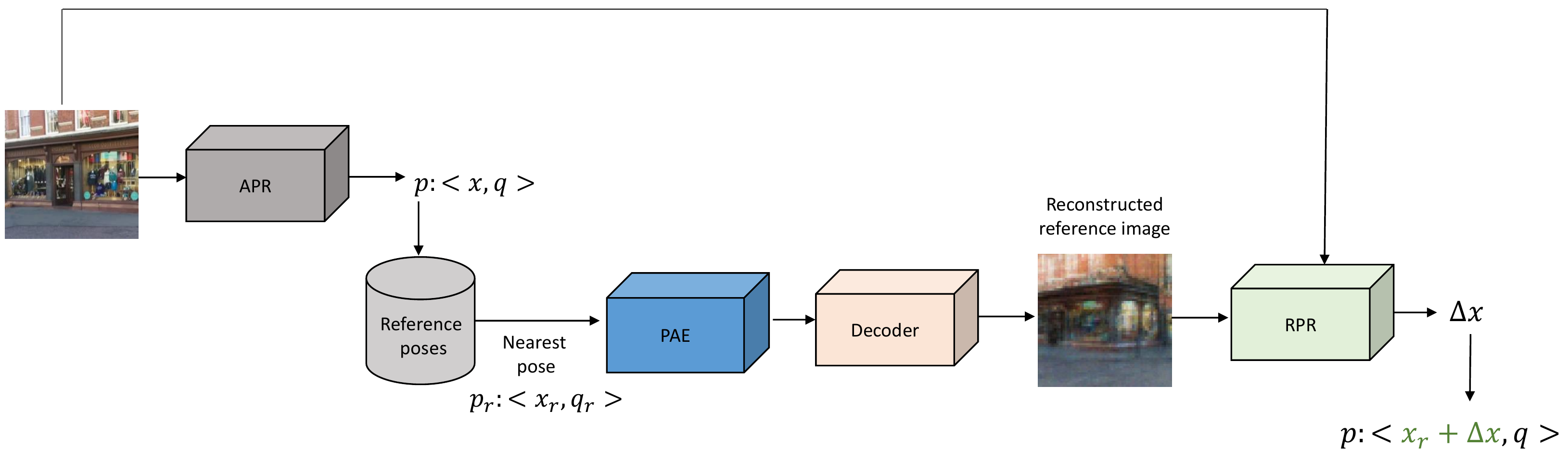}
\caption{Virtual relative pose regression for position estimation.}
\label{fig:virtual-rpr}
\vspace{-2em}
\end{figure}

The proposed pose embedding encodes both
visual and geometric information, allowing to reconstruct the
respective image given \textit{only} the input pose $\mathbf{p:<x,q>}$. This
can be achieved by training a simple MLP decoder $\mathbf{D}$ to minimize
the $\mathbb{L}_{1}$ loss between the original and reconstructed images, as
illustrated in Fig. \ref{fig:img-recon}.
The ability to reconstruct images from pose encoding paves the way for
performing \textit{virtual} relative pose regression. While in regression-based RPR, the images are encoded by a CNN, we propose to encode only the localization parameters using the PAE. Specifically, as opposed to common relative
pose regression, where the relative motion is regressed from latent image
encoding of the query and nearest images, here we can encode reconstructed
images 'on-the-fly'. We can further exploit the \textit{virtual} pose regression
to improve the localization of APR (Fig. \ref{fig:virtual-rpr}). Similarly to
our test-time optimization procedure, we start by computing the pose
estimate $\mathbf{p:<x,q>}$ from the query image using an APR $\mathbf{A}$.
We then retrieve the closest train reference pose, encode it with a
pre-trained pose auto-encoder $\mathbf{f}$ and reconstruct the image with a
pre-trained decoder $\mathbf{D}$. Given the query image and the reconstructed
train image, a pretrained RPR can be applied to regress
the relative translation from which a refined position estimate can be
obtained.

\subsection{Implementation Details}

\label{subsubsec:Implementation Details}The proposed PAE consists of two MLP
heads, each with four fully connected (FC) layers with ReLU non-linearity,
expanding the initial Fourier Feature dimension to 64, 128, 256 and $d$,
the APR latent dimension, respectively. In our experiments, we set $d=256$,
for all APR architectures. We apply Eq.~\ref{eq:fourier_features} with $L=6$
for encoding $x$, $q$ as well as the scene index $s$ for multiscene PAEs.
For training and evaluation, we consider different single- and multi- scene
APR teachers: a PoseNet-like \cite{kendall2015posenet} architecture with
different convolutional backbones (MobileNet\cite{howard2017mobilenets},
ResNet50\cite{he2016deep} and EfficientNet-B0 \cite{pmlr-v97-tan19a}), and a
recent state-of-the-art transformer-based APR (MS-Transformer\cite%
{ShavitFerensIccv21}). We implement PoseNet-like APRs using a convolutional
backbone of choice and an additional two FC layers and ReLU nonlinearity to
map the backbone dimension to $d$ and generate the respective latent
representations for $\textbf{x}$ and $\textbf{q}$. The regressor head consist of two FC layers
to regress $\textbf{x}$ and $\textbf{q}$, respectively. For MS-Transformer, we used the
pretrained implementation provided by the authors. Our test-time
optimization is implemented with $k=3$ nearest neighbors and $n=3$
iterations. For image reconstruction we use a four-layer MLP decoder with
ReLU non-linearity, increasing the initial encoding dimension $d$ to 512,
1024, 2048 and $3hw^{2}$, where $h$ and $w$, the height and width of the
reconstructed image, are set to 64. In order to perform virtual relative
pose regression, we apply a Siamese network with a similar architecture to
our PoseNet-like APRs. We use Efficient-B0 for the convolutional backbone
and apply it twice. The resulting flattened activation maps are concatenated
and then used to regress $\textbf{x}$ and $\textbf{q}$ as in PoseNet-like APRs (the only
difference is in the first FC layer, which maps from twice the backbone
dimension to $d$). We implement all the models and the proposed procedures in
PyTorch \cite{paszke2019pytorch}. Training and inference were performed on
an NVIDIA GeForce GTX 1080 GPU with 8Gb. In order to support easy
reproduction of the reported results, we provide the implementation of all
the architectures and procedures described in this paper and make our code
and pre-trained models publicly available.

\section{Experimental Results}

\label{sec:Experimental Results}

\subsection{Experimental Setup}

\label{subsec:Experimental Setup} \textbf{Datasets. }The proposed PAE scheme
is evaluated using the 7Scenes \cite{glocker2013real} and the Cambridge
Landmarks \cite{kendall2015posenet} datasets, which are commonly benchmarked
in contemporary pose regression works \cite%
{kendall2015posenet,kendall2017geometric,ShavitFerensIccv21}. The 7Scenes
dataset consists of seven small-scale scenes ($\sim 1-10m^{2}$) depicting an
indoor office environment. There are six scenes in the Cambridge Landmark
dataset \textbf{\ ($\sim 900-5500m^{2}$) } captured at outdoor urban locations, out of
which four scenes were considered for our comparative analysis as they are
typically used for evaluating APRs.

\textbf{Training Details. }We optimize the single-scene APR teachers using
Adam, with $\beta _{1}=0.9$, $\beta _{2}=0.999$ and $\epsilon =10^{-10}$. We
minimize the learned pose loss (Eq. \ref{equ:learnable pose loss}) and
initialize its parameters as in \cite{valada2018deep}. Each APR is trained
for 300 epochs, with a batch size of $32$ and an initial learning rate of $%
10^{-3}$. For the MS-Transformer teacher, we use the provided pretrained
models \cite{ShavitFerensIccv21} for the CambridgeLandmarks and 7Scenes datasets. The PAEs are trained
using the same training configuration as their teachers when optimizing the
loss in Eq. \ref{equ:pose encoder loss}. Our test-time optimization is
performed with AdamW and a learning rate of $10^{-3}$.
We applied Adam to optimize our decoder and relative pose regressor, with
initial learning rates of $10^{-2}$ and $10^{-3}$, respectively. Additional
augmentation and training details are provided in the supplementary materials (suppl. materials).
\subsection{Evaluation of Camera Pose Auto-Encoders (PAEs)}
\label{subsec:PAE evaluation}We evaluate the proposed PAEs by comparing the
original localization error of the teacher APR and the error observed when
using the APR's head to regress the pose from the PAE encoding. We report
the results for the CambridgeLandmarks (Table \ref{tb:cambridge_decoding})
and 7Scenes (Table \ref{tb:7scenes_decoding}) datasets, respectively, using
the MS-Transformer as the teacher APR. \vspace{-1em}
\begin{table}[tbh]
\caption{Median position/orientation error in meters/degrees, when learning
from images and when decoding a latent pose encoding from a student PAE. We
use MS-Transformer \cite{ShavitFerensIccv21}, pre-trained on the
CambridgeLandmarks dataset, as our teacher APR.}
\label{tb:cambridge_decoding}\centering\setlength{\tabcolsep}{4pt} {\small
\begin{tabular}{cccccc}
\hline
\textbf{Method} & \textbf{K. College} & \textbf{Old Hospital} & \textbf{Shop
Facade} & \textbf{St. Mary} &  \\ \hline
\multicolumn{1}{l}{Teacher APR} & {0.83}/{1.47} & {1.81}/{2.39} & {0.86}/{%
3.07} & {1.62}/ {3.99} &  \\ \hline
\multicolumn{1}{l}{Student PAE} & {0.90}/{1.49} & {2.07}/{2.58} & {0.99}/{%
3.88} & {1.64}/ {4.16} &  \\ \hline
\end{tabular}%
} \vspace{-1em}
\end{table}
The student auto-encoder obtains an accuracy similar to the teacher
APR, across both datasets. While in most cases, the student accuracy is
still inferior with respect to the teacher, in some cases (e.g., the
orientation error for the Fire scene), the student provides a better
estimation. \vspace{-1em}
\begin{table}[tbh]
\caption{Median position/orientation error in meters/degrees, when learning from
images and when decoding a latent pose encoding from a student PAE (S. PAE).
We use MS-Transformer\protect\cite{ShavitFerensIccv21}, pre-trained on the
7Scenes dataset, as our teacher APR (T. APR).}
\label{tb:7scenes_decoding}\setlength{\tabcolsep}{2pt} \centering
{\small
\begin{tabular}{cccccccc}
\hline
\textbf{Method} & \textbf{Chess} & \textbf{Fire} & \textbf{Heads} & \textbf{%
Office} & \textbf{Pumpkin} & \textbf{Kitchen} & \textbf{Stairs} \\ \hline
\multicolumn{1}{l}{T. APR} & 0.11/{4.66} & {0.24 }/{9.60} & {0.14}/{12.2} &
0.17/{5.66} & {0.18}/{4.44} & {0.17}/{5.94} & {0.26}/{8.45} \\
\multicolumn{1}{l}{S. PAE} & 0.12/{4.95} & {0.24}/ {9.31} & {0.14}/{12.5} &
0.19/{5.79} & {0.18}/{4.89} & {0.18}/{6.19} & {0.25}/{8.74} \\ \hline
\end{tabular}%
} \vspace{-2em}
\end{table}

\subsection{Ablation Study}

\label{subsec:Ablation Study}We further carry out different ablations to
assess the proposed PAE architecture and the robustness of the
proposed concept in different teacher APRs. Table~\ref{tb:ablation-arch}
shows the median position and orientation errors for the KingsCollege scene from the CambridgeLandmarks dataset,
obtained with three different PAE architectures: 2-layers MLP, 4-layers MLP
and a 4-layers MLP applied in conjunction with Fourier Features (selected architecture). Although all
three variants achieve similar performance, the latter achieves the best
trade-off between position and orientation. Additional ablation study of the dimensionality of Fourier Features (the effect of $L$)  is provided in our suppl. materials (suppl. section 1.3). \vspace{-1em}
\begin{table}[tbh]
\caption{Ablations of the PAE architecture. We compare the
median position and orientation errors when using shallow and deep MLP
architectures with and without Fourier Features (position encoding). The performance is reported for the KingsCollege scene (CambridgeLandmarks dataset). The
Teacher is a PoseNet APR with a MobileNet architecture.}
\label{tb:ablation-arch}\centering{\small
\begin{tabular}{lll}
\hline
\textbf{Auto Encoder Architecture} & \textbf{Position [m]} & \textbf{%
Orientation [deg]} \\ \hline
2-Layers MLP & \multicolumn{1}{c}{1.27} & \multicolumn{1}{c}{\textbf{3.41}}
\\
4-Layers MLP & \multicolumn{1}{c}{1.26} & \multicolumn{1}{c}{3.54} \\
Fourier Features + 4-Layers MLP & \multicolumn{1}{c}{\textbf{1.15}} &
\multicolumn{1}{c}{3.58} \\ \hline
\end{tabular}%
} \vspace{-1em}
\end{table}

Since PAEs are not limited to a particular APR teacher,
we further evaluate several single- and multi- scene APR teacher architectures:
three PoseNet variants with different convolutional backbones and
MS-Transformer. Table \ref{table:ablation-apr} shows the results for the
KingsCollege scene. The student
auto-encoder is able to closely reproduce its teacher's performance,
regardless of the specific architecture used. \vspace{-1em}
\begin{table}[tbh]
\caption{Ablations of the teacher (single/multi-scene) APR architecture. We
compare the median position and orientation errors when training on images
and when decoding from a student auto-encoder. The performance is reported for
the KingsCollege scene (CambridgeLandmarks dataset).}
\label{table:ablation-apr}\centering{\small
\begin{tabular}{lcc}
\hline
\textbf{APR Architecture} & \textbf{Teacher APR} & \textbf{Student
PAE} \\
& [m/deg] & [m/deg] \\ \hline
PoseNet+MobileNet & 1.24/3.45 & 1.15/3.58 \\
PoseNet+ResNet50 & 1.56/3.79 & 1.50/3.77 \\
PoseNet+EfficientNet & 0.88/2.91 & \textbf{0.83}/2.97 \\
MS-Transformer & \textbf{0.83/1.47} & 0.90/\textbf{1.49} \\ \hline
\end{tabular}%
}\vspace{-1em}
\end{table}

Learning to encode camera poses allows us to leverage available prior
information at a potentially low cost. We report the runtime and memory
requirements associated with using a PAE and with retrieving and storing
reference poses (Table \ref{tb:requirements}). Applying a multi-scene PAE
requires an additional runtime of $1.22$ms and $<1$Mb for the model's
weights. Storing all poses from the CambridgeLandmarks and 7Scenes
datasets incurs a total of $2.15$Mb with an average retrieval runtime of $0.16
$ms.\vspace{-1.5em}
\begin{table}[tbh]
\caption{Additional runtime and memory required for using a
PAE, and retrieving and storing reference poses.}
\label{tb:requirements}\centering{\small
\begin{tabular}{lcc}
\hline
\textbf{Requirement} & \textbf{Runtime} & \textbf{Memory} \\
& \textbf{\ [ms]} & \textbf{\ [Mb] } \\ \hline
Components &  &  \\ \hline
Camera Pose Auto-Encoder & 1.22 & 0.89 \\
Retrieving and Storing Poses & 0.16 & 2.15 \\ \hline
\end{tabular}%
}
\end{table}
\vspace{-3em}
\subsection{Refining Position Estimation with Encoded Poses}
\label{subsec:exp refine}We evaluate the proposed use of PAEs (section \ref%
{subsec:applications}) for position refinement and image reconstruction.
Tables ~\ref{tb:cambridge_rank} and \ref{tb:7scenes_rank} show the average
of median position/orientation errors in meters/degrees obtained for the
CambridgeLandmarks and 7Scenes datasets, respectively. We report the results
of single-scene and multi-scene APRs and the result when refining the
position with our test-time optimization procedure for MS-Transformer (orientation is estimated with MS-
Transformer without refinement). Using camera pose encoding of the train images achieves a new SOTA accuracy for
absolute pose regression on both datasets. Specifically, we improve the
average position error of the current SOTA APR (MS-Transformer) from $1.28$
meter to a sub-meter error ($0.96$ meters) for the CambridgeLandmarks
dataset and reduce it by $17\%$ for the 7Scenes dataset ($0.15$ versus $018$%
, respectively). We report additional results for single-scene APRs with position refinement as well as verification results obtained when starting from an initial guess of the pose, sampled around
the ground truth pose, in our suppl. materials (suppl. section 1.4). Our test-time optimization achieves a consistent trend of improvement regardless of the specific APR architecture used and across scenes and datasets. The total additional runtime required for the proposed test-time
optimization (retrieving poses, encoding them and computing the weights of
the affine transformation) is $7.51$ms.

We further explore the application of camera pose encoding for image
reconstruction and virtual relative pose regression. Fig. \ref%
{fig:img-recon1} shows the original and reconstructed images from the Shop
Facade (Cambridge Landmarks dataset) and the Heads (7Scenes dataset) scenes.
Our simple MLP decoder learns to decode images at a 64x64 resolution. Although
the reconstructed images are blurry, their main visually identifying
features are clearly visible. In the context of our work, image reconstruction aims to serve virtual
relative pose regression for refining the position of APRs. Table \ref%
{table:ror} reports the median position error for the ShopFacade and Heads
scenes, for single scene and multi-scene APRs, and when refining the
position through image reconstruction and relative pose regression (section %
\ref{subsec:applications}). For both scenes, the proposed procedure improves
the position accuracy of the teacher APR's initial estimation and achieves a
new SOTA position accuracy for absolute pose regression. The
total run time required for this procedure (retrieving the closest pose,
encoding it, decoding the image, applying the regressor, and computing the new position) is $15.31$ms.
\begin{table}[th!]
\caption{Localization results for the Cambridge Landmarks dataset. We report
the average of median position/orientation errors in meters/degrees. The best
results are highlighted in bold.}
\label{tb:cambridge_rank}\centering
{\small
\begin{tabular}{cc}
\hline
\textbf{APR Architecture} & \textbf{Average} \textbf{[m/deg]} \\ \hline
\multicolumn{1}{l}{PoseNet \cite{kendall2015posenet}} & 2.09/6.84 \\
\multicolumn{1}{l}{BayesianPN \cite{kendall2016modelling}} & 1.92/6.28 \\
\multicolumn{1}{l}{LSTM-PN \cite{walch2017image}} & 1.30/5.52 \\
\multicolumn{1}{l}{SVS-Pose \cite{naseer2017deep}} & 1.33/5.17 \\
\multicolumn{1}{l}{GPoseNet \cite{cai2019hybrid}} & 2.08/4.59 \\
\multicolumn{1}{l}{PoseNet-Learnable \cite{kendall2017geometric}} & 1.43/2.85
\\
\multicolumn{1}{l}{GeoPoseNet \cite{kendall2017geometric}} & 1.63/2.86 \\
\multicolumn{1}{l}{MapNet \cite{brahmbhatt2018geometry}} & 1.63/3.64 \\
\multicolumn{1}{l}{IRPNet \cite{shavitferensirpnet}} & 1.42/3.45 \\
\multicolumn{1}{l}{MSPN \cite{blanton2020extending}} & 2.47/5.34 \\
\multicolumn{1}{l}{MS-Transformer \cite{ShavitFerensIccv21}} & 1.28/\textbf{%
2.73 } \\
\multicolumn{1}{l}{\textbf{MS-Transformer + Optimized Position (Ours)}} &
\textbf{0.96}/\textbf{2.73} \\ \hline
\end{tabular}%
} \vspace{-0em}
\end{table}
\begin{table}[tbh]
\caption{Localization results for the 7Scenes dataset. We report the average
of median position/orientation errors in meters/degrees. The best results are
highlighted in bold. }
\label{tb:7scenes_rank}\centering{\small
\begin{tabular}{cc}
\hline
\textbf{APR Architecture} & \textbf{Average} \textbf{[m/deg]} \\ \hline
\multicolumn{1}{l}{PoseNet \cite{kendall2015posenet}} & 0.44/10.4 \\
\multicolumn{1}{l}{BayesianPN \cite{kendall2016modelling}} & 0.47/9.81 \\
\multicolumn{1}{l}{LSTM-PN \cite{walch2017image}} & 0.31/9.86 \\
\multicolumn{1}{l}{GPoseNet \cite{cai2019hybrid}} & 0.31/9.95 \\
\multicolumn{1}{l}{PoseNet-Learnable \cite{kendall2017geometric}} & 0.24/7.87
\\
\multicolumn{1}{l}{GeoPoseNet \cite{kendall2017geometric}} & 0.23/8.12 \\
\multicolumn{1}{l}{MapNet \cite{brahmbhatt2018geometry}} & 0.21/7.78 \\
\multicolumn{1}{l}{IRPNet \cite{shavitferensirpnet}} & 0.23/8.49 \\
\multicolumn{1}{l}{AttLoc \cite{wang2020atloc}} & 0.20/7.56 \\
\multicolumn{1}{l}{MSPN \cite{blanton2020extending}} & 0.20/8.41 \\
\multicolumn{1}{l}{MS-Transformer \cite{ShavitFerensIccv21}} & {0.18}/%
\textbf{\ {7.28}} \\
\multicolumn{1}{l}{\textbf{MS-Transformer+Optimized Position (Ours)}} &
\textbf{0.15}/ \textbf{7.28} \\ \hline
\end{tabular}%
} \vspace{-2em}
\end{table}
\begin{figure}[th!]
\begin{center}
\begin{tabular}{ccc}
\subfigure{\includegraphics[width=0.15\linewidth]{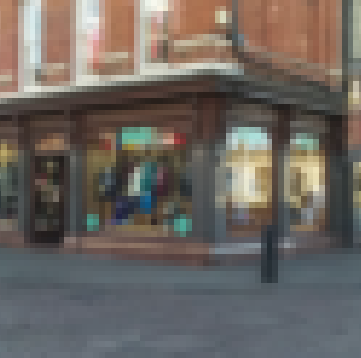}}
& \subfigure{\includegraphics[width=0.15\linewidth]{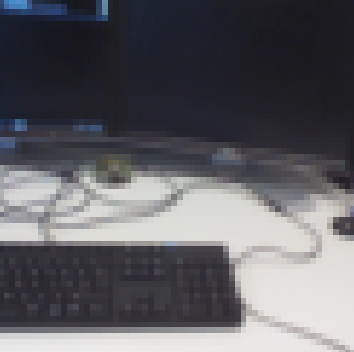}} & \subfigure{\includegraphics[width=0.15\linewidth]{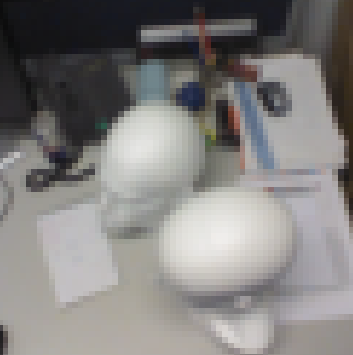}} \\
(a) & (b) & (c) \\
\subfigure{\includegraphics[width=0.15\linewidth]{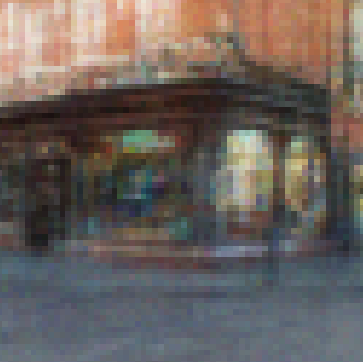}}
& \subfigure{\includegraphics[width=0.15
\linewidth]{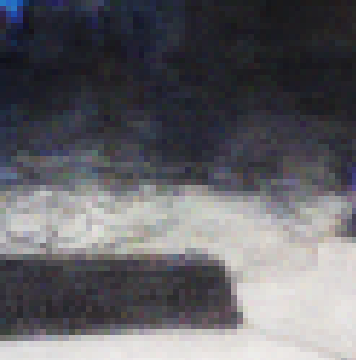}} & \subfigure{%
\includegraphics[width=0.15\linewidth]{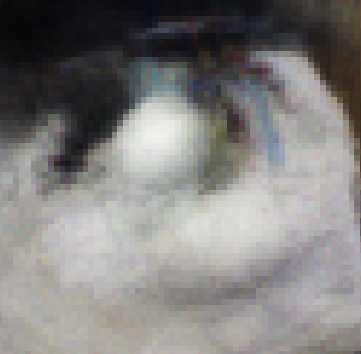}}\\
(d) & (e) & (f)
\end{tabular}
\end{center}
\vspace{-1em} \vspace{-1em}
\caption{Images reconstructed from learned camera pose encoding. (a)-(c)
Original images from the Shop Facade and Heads scenes at a 64x64 resolution.
(d)-(e) Corresponding reconstructed images.}
\label{fig:img-recon1}
\end{figure} \vspace{-2em}
\subsection{Limitations and Future Research}
Although our work demonstrates useful applications of the proposed PAEs for
advancing APR accuracy, they focus on position estimation and image
reconstruction. Our preliminary experiments show that for orientation
estimation, the proposed encoding can provide a reasonable estimate but does
not advance SOTA APR accuracy (suppl. section 1.5). Further research into
orientation-optimized encoding, as well as different architecture choices
for our decoder and relative pose regressor, are directions for further
improvements. Another interesting aspect is the ability of camera PAEs to increase the resolution of the training set by encoding
virtual unseen poses, which can enrich existing datasets with
a minimal cost. We also note that APRs are a family of methods within a larger body
of localization works (section \ref{sec:related}). Although our work focuses on
advancing the accuracy of APRs and extending them to use prior information, while maintaining its advantages (lightweight, fast, and robust to query
camera intrinsics), it is still inferior to structure-based methods
in terms of accuracy. We provide a comparison of different representative
localization schemes to show the current gaps and advancements made (suppl. section 1.6).
\vspace{-1.8em}
\begin{table}[th!]
\caption{Median position error with/without virtual relative pose regression
for the ShopFacdade and Heads scenes (orientation error remains fixed).}
\label{table:ror}\centering{\small
\begin{tabular}{lcc}
\hline
\textbf{APR Architecture} & \textbf{Shop Facade} & \textbf{Heads} \\
\multicolumn{1}{c}{} & [m] & [m] \\ \hline
PoseNet \cite{kendall2015posenet} & 1.46 & 0.29 \\
BayesianPN \cite{kendall2016modelling} & 1.25 & 0.31 \\
LSTM-PN \cite{walch2017image} & 1.18 & 0.21 \\
SVS-Pose \cite{svs_pose} & 0.63 & $--$ \\
GPoseNet \cite{cai2019hybrid} & 1.14 & 0.21 \\
PoseNet-Learnable \cite{kendall2017geometric} & 1.05 & 0.18 \\
GeoPoseNet \cite{kendall2017geometric} & 0.88 & 0.17 \\
MapNet \cite{brahmbhatt2018geometry} & 1.49 & 0.18 \\
IRPNet \cite{shavitferensirpnet} & 0.72 & 0.15 \\
AttLoc \cite{wang2020atloc} & $--$ & 0.61 \\
MSPN \cite{blanton2020extending} & 2.92 & 0.16 \\
MS-Transformer \cite{shavit2019introduction} & {0.86} & 0.14 \\
MS-Transformer + Virtual RPR (ours) & {\textbf{0.62}} & {\textbf{0.10}} \\
\hline
\end{tabular}%
}
\end{table}
\vspace{-2.5em}
\section{Conclusions}
In this paper, we proposed Camera Pose Auto-Encoders for encoding camera
poses into latent representations that can be used for absolute and relative
pose regression. Encoding camera poses paves the way for introducing visual
and geometric priors with relatively minor runtime and memory costs, and is
shown to improve position estimation and achieve a new SOTA absolute pose
regression accuracy across contemporary outdoor and indoor benchmarks.

\clearpage
%
%
\bibliographystyle{splncs04}
\bibliography{pae}

\begin{thebibliography}{10}
\providecommand{\url}[1]{\texttt{#1}}
\providecommand{\urlprefix}{URL }
\providecommand{\doi}[1]{https://doi.org/#1}

\bibitem{balntas2018relocnet}
Balntas, V., Li, S., Prisacariu, V.: Relocnet: Continuous metric learning
  relocalisation using neural nets. In: Proceedings of the European Conference
  on Computer Vision (ECCV) (September 2018)

\bibitem{blanton2020extending}
Blanton, H., Greenwell, C., Workman, S., Jacobs, N.: Extending absolute pose
  regression to multiple scenes. In: Proceedings of the IEEE/CVF Conference on
  Computer Vision and Pattern Recognition Workshops. pp. 38--39 (2020)

\bibitem{DSAC}
Brachmann, E., Krull, A., Nowozin, S., Shotton, J., Michel, F., Gumhold, S.,
  Rother, C.: Dsac - differentiable ransac for camera localization. In: 2017
  IEEE Conference on Computer Vision and Pattern Recognition (CVPR). pp.
  2492--2500. IEEE Computer Society, Los Alamitos, CA, USA (jul 2017).
  \doi{10.1109/CVPR.2017.267},
  \url{https://doi.ieeecomputersociety.org/10.1109/CVPR.2017.267}

\bibitem{DSAC++}
{Brachmann}, E., {Rother}, C.: Learning less is more - 6d camera localization
  via 3d surface regression. In: 2018 IEEE/CVF Conference on Computer Vision
  and Pattern Recognition. pp. 4654--4662 (2018). \doi{10.1109/CVPR.2018.00489}

\bibitem{9394752}
Brachmann, E., Rother, C.: Visual camera re-localization from rgb and rgb-d
  images using dsac. IEEE Transactions on Pattern Analysis and Machine
  Intelligence (01), ~1--1 (apr 2021)

\bibitem{brahmbhatt2018geometry}
Brahmbhatt, S., Gu, J., Kim, K., Hays, J., Kautz, J.: Geometry-aware learning
  of maps for camera localization. In: IEEE Conference on Computer Vision and
  Pattern Recognition (CVPR) (2018)

\bibitem{cai2019hybrid}
Cai, M., Shen, C., Reid, I.: A hybrid probabilistic model for camera
  relocalization  (2019)

\bibitem{DBLP:conf/cvpr/CavallariGLVST17}
Cavallari, T., Golodetz, S., Lord, N.A., Valentin, J.P.C., di~Stefano, L.,
  Torr, P.H.S.: On-the-fly adaptation of regression forests for online camera
  relocalisation. In: 2017 {IEEE} Conference on Computer Vision and Pattern
  Recognition, {CVPR} 2017, Honolulu, HI, USA, July 21-26, 2017. pp. 218--227.
  {IEEE} Computer Society (2017)

\bibitem{ding2019camnet}
Ding, M., Wang, Z., Sun, J., Shi, J., Luo, P.: Camnet: Coarse-to-fine retrieval
  for camera re-localization. In: Proceedings of the IEEE/CVF International
  Conference on Computer Vision (ICCV) (October 2019)

\bibitem{dusmanu2019d2}
{Dusmanu}, M., {Rocco}, I., {Pajdla}, T., {Pollefeys}, M., {Sivic}, J.,
  {Torii}, A., {Sattler}, T.: D2-net: A trainable cnn for joint description and
  detection of local features. In: 2019 IEEE/CVF Conference on Computer Vision
  and Pattern Recognition (CVPR). pp. 8084--8093 (2019).
  \doi{10.1109/CVPR.2019.00828}

\bibitem{fischler1981random}
Fischler, M.A., Bolles, R.C.: Random sample consensus: A paradigm for model
  fitting with applications to image analysis and automated cartography.
  Commun. ACM  \textbf{24}(6),  381–395 (Jun 1981)

\bibitem{glocker2013real}
{Glocker}, B., {Izadi}, S., {Shotton}, J., {Criminisi}, A.: Real-time rgb-d
  camera relocalization. In: 2013 IEEE International Symposium on Mixed and
  Augmented Reality (ISMAR). pp. 173--179 (2013).
  \doi{10.1109/ISMAR.2013.6671777}

\bibitem{he2016deep}
{He}, K., {Zhang}, X., {Ren}, S., {Sun}, J.: Deep residual learning for image
  recognition. In: 2016 IEEE Conference on Computer Vision and Pattern
  Recognition (CVPR). pp. 770--778 (2016). \doi{10.1109/CVPR.2016.90}

\bibitem{howard2017mobilenets}
Howard, A.G., Zhu, M., Chen, B., Kalenichenko, D., Wang, W., Weyand, T.,
  Andreetto, M., Adam, H.: Mobilenets: Efficient convolutional neural networks
  for mobile vision applications. arXiv preprint arXiv:1704.04861  (2017)

\bibitem{kendall2017geometric}
{Kendall}, A., {Cipolla}, R.: Geometric loss functions for camera pose
  regression with deep learning. In: 2017 IEEE Conference on Computer Vision
  and Pattern Recognition (CVPR). pp. 6555--6564 (2017).
  \doi{10.1109/CVPR.2017.694}

\bibitem{kendall2015posenet}
{Kendall}, A., {Grimes}, M., {Cipolla}, R.: Posenet: A convolutional network
  for real-time 6-{DOF} camera relocalization. In: 2015 IEEE International
  Conference on Computer Vision (ICCV). pp. 2938--2946 (2015).
  \doi{10.1109/ICCV.2015.336}

\bibitem{kendall2016modelling}
Kendall, A., Cipolla, R.: Modelling uncertainty in deep learning for camera
  relocalization. In: Proceedings of the International Conference on Robotics
  and Automation ({ICRA}) (2016)

\bibitem{melekhov2017image}
Melekhov, I., Ylioinas, J., Kannala, J., Rahtu, E.: Image-based localization
  using hourglass networks. In: 2017 {IEEE} International Conference on
  Computer Vision Workshops, {ICCV} Workshops 2017, Venice, Italy, October
  22-29, 2017. pp. 870--877. {IEEE} Computer Society (2017).
  \doi{10.1109/ICCVW.2017.107}

\bibitem{Trujillo}
Mera-Trujillo, M., Smith, B., Fragoso, V.: Efficient scene compression for
  visual-based localization. In: 2020 International Conference on 3D Vision
  (3DV). pp. 1--10. IEEE Computer Society, Los Alamitos, CA, USA (nov 2020).
  \doi{10.1109/3DV50981.2020.00111},
  \url{https://doi.ieeecomputersociety.org/10.1109/3DV50981.2020.00111}

\bibitem{mildenhall2020nerf}
Mildenhall, B., Srinivasan, P.P., Tancik, M., Barron, J.T., Ramamoorthi, R.,
  Ng, R.: Nerf: Representing scenes as neural radiance fields for view
  synthesis. In: European conference on computer vision. pp. 405--421. Springer
  (2020)

\bibitem{naseer2017deep}
Naseer, T., Burgard, W.: Deep regression for monocular camera-based 6-{DoF}
  global localization in outdoor environments. 2017 IEEE/RSJ International
  Conference on Intelligent Robots and Systems (IROS) pp. 1525--1530 (2017)

\bibitem{svs_pose}
Naseer, T., Burgard, W.: Deep regression for monocular camera-based {6-DoF}
  global localization in outdoor environments. In: IROS (2017)

\bibitem{noh2017large}
{Noh}, H., {Araujo}, A., {Sim}, J., {Weyand}, T., {Han}, B.: Large-scale image
  retrieval with attentive deep local features. In: 2017 IEEE International
  Conference on Computer Vision (ICCV). pp. 3476--3485 (2017).
  \doi{10.1109/ICCV.2017.374}

\bibitem{paszke2019pytorch}
Paszke, A., Gross, S., Massa, F., Lerer, A., Bradbury, J., Chanan, G., Killeen,
  T., Lin, Z., Gimelshein, N., Antiga, L., Desmaison, A., Kopf, A., Yang, E.,
  DeVito, Z., Raison, M., Tejani, A., Chilamkurthy, S., Steiner, B., Fang, L.,
  Bai, J., Chintala, S.: Pytorch: An imperative style, high-performance deep
  learning library. In: Wallach, H., Larochelle, H., Beygelzimer, A.,
  Alche-Buc, F., Fox, E., Garnett, R. (eds.) Advances in Neural Information
  Processing Systems. vol.~32, pp. 8026--8037. Curran Associates, Inc. (2019)

\bibitem{radwan2018vlocnet++}
{Radwan}, N., {Valada}, A., {Burgard}, W.: Vlocnet++: Deep multitask learning
  for semantic visual localization and odometry. IEEE Robotics and Automation
  Letters  \textbf{3}(4),  4407--4414 (2018). \doi{10.1109/LRA.2018.2869640}

\bibitem{rahaman2018spectral}
Rahaman, N., Arpit, D., Baratin, A., Draxler, F., Lin, M., Hamprecht, F.A.,
  Bengio, Y., Courville, A.C.: On the spectral bias of deep neural networks.
  (2018)

\bibitem{DBLP:conf/bmvc/SahaVJ18}
Saha, S., Varma, G., Jawahar, C.V.: Improved visual relocalization by
  discovering anchor points. In: British Machine Vision Conference 2018, {BMVC}
  2018, Newcastle, UK, September 3-6, 2018. p.~164. {BMVA} Press (2018)

\bibitem{sarlin2019coarse}
{Sarlin}, P., {Cadena}, C., {Siegwart}, R., {Dymczyk}, M.: From coarse to fine:
  Robust hierarchical localization at large scale. In: 2019 IEEE/CVF Conference
  on Computer Vision and Pattern Recognition (CVPR). pp. 12708--12717 (2019).
  \doi{10.1109/CVPR.2019.01300}

\bibitem{sarlin21pixloc}
Sarlin, P.E., Unagar, A., Larsson, M., Germain, H., Toft, C., Larsson, V.,
  Pollefeys, M., Lepetit, V., Hammarstrand, L., Kahl, F., Sattler, T.: {Back to
  the Feature: Learning Robust Camera Localization from Pixels to Pose}. In:
  CVPR (2021)

\bibitem{sattler2016efficient}
{Sattler}, T., {Leibe}, B., {Kobbelt}, L.: Efficient \& effective prioritized
  matching for large-scale image-based localization. IEEE Transactions on
  Pattern Analysis and Machine Intelligence  \textbf{39}(9),  1744--1756
  (2017). \doi{10.1109/TPAMI.2016.2611662}

\bibitem{sattler2019understanding}
{Sattler}, T., {Zhou}, Q., {Pollefeys}, M., {Leal-Taixé}, L.: Understanding
  the limitations of cnn-based absolute camera pose regression. In: 2019
  IEEE/CVF Conference on Computer Vision and Pattern Recognition (CVPR). pp.
  3297--3307 (2019). \doi{10.1109/CVPR.2019.00342}

\bibitem{shavit2019introduction}
Shavit, Y., Ferens, R.: Introduction to camera pose estimation with deep
  learning (2019)

\bibitem{shavitferensirpnet}
Shavit, Y., Ferens, R.: Do we really need scene-specific pose encoders. In: To
  Appear in 2021 IEEE International Conference on Pattern Recognition (ICPR)
  (2021)

\bibitem{ShavitFerensIccv21}
Shavit, Y., Ferens, R., Keller, Y.: Learning multi-scene absolute pose
  regression with transformers. In: 2021 IEEE International Conference on
  Computer Vision (ICCV) (2021)

\bibitem{shotton2013scene}
Shotton, J., Glocker, B., Zach, C., Izadi, S., Criminisi, A., Fitzgibbon, A.:
  Scene coordinate regression forests for camera relocalization in rgb-d
  images. In: Proc. Computer Vision and Pattern Recognition (CVPR). IEEE (June
  2013)

\bibitem{taira2018inloc}
{Taira}, H., {Okutomi}, M., {Sattler}, T., {Cimpoi}, M., {Pollefeys}, M.,
  {Sivic}, J., {Pajdla}, T., {Torii}, A.: Inloc: Indoor visual localization
  with dense matching and view synthesis. IEEE Transactions on Pattern Analysis
  and Machine Intelligence pp.~1--1 (2019). \doi{10.1109/TPAMI.2019.2952114}

\bibitem{pmlr-v97-tan19a}
Tan, M., Le, Q.: {E}fficient{N}et: Rethinking model scaling for convolutional
  neural networks. Proceedings of Machine Learning Research, vol.~97, pp.
  6105--6114. PMLR, Long Beach, California, USA (09--15 Jun 2019)

\bibitem{tancik2020fourier}
Tancik, M., Srinivasan, P., Mildenhall, B., Fridovich-Keil, S., Raghavan, N.,
  Singhal, U., Ramamoorthi, R., Barron, J., Ng, R.: Fourier features let
  networks learn high frequency functions in low dimensional domains. Advances
  in Neural Information Processing Systems  \textbf{33},  7537--7547 (2020)

\bibitem{Torii}
Torii, A., Arandjelovic, R., Sivic, J., Okutomi, M., Pajdla, T.: 24/7 place
  recognition by view synthesis. IEEE Trans. Pattern Anal. Mach. Intell.
  \textbf{40}(2),  257–271 (feb 2018)

\bibitem{9665967}
Turkoglu, M., Brachmann, E., Schindler, K., Brostow, G.J., Monszpart, A.:
  Visual camera re-localization using graph neural networks and relative pose
  supervision. In: 2021 International Conference on 3D Vision (3DV). pp.
  145--155. Los Alamitos, CA, USA (dec 2021)

\bibitem{valada2018deep}
Valada, A., Radwan, N., Burgard, W.: Deep auxiliary learning for visual
  localization and odometry. ICRA pp. 6939--6946 (2018)

\bibitem{walch2017image}
{Walch}, F., {Hazirbas}, C., {Leal-Taixé}, L., {Sattler}, T., {Hilsenbeck},
  S., {Cremers}, D.: Image-based localization using lstms for structured
  feature correlation. In: 2017 IEEE International Conference on Computer
  Vision (ICCV). pp. 627--637 (2017). \doi{10.1109/ICCV.2017.75}

\bibitem{wang2020atloc}
Wang, B., Chen, C., Lu, C.X., Zhao, P., Trigoni, N., Markham, A.: Atloc:
  Attention guided camera localization. In: Proceedings of the AAAI Conference
  on Artificial Intelligence. vol.~34, pp. 10393--10401 (2020)

\bibitem{wu2017delving}
{Wu}, J., {Ma}, L., {Hu}, X.: Delving deeper into convolutional neural networks
  for camera relocalization. In: 2017 IEEE International Conference on Robotics
  and Automation (ICRA). pp. 5644--5651 (2017). \doi{10.1109/ICRA.2017.7989663}

\bibitem{9156582}
Xue, F., Wu, X., Cai, S., Wang, J.: Learning multi-view camera relocalization
  with graph neural networks. In: 2020 IEEE/CVF Conference on Computer Vision
  and Pattern Recognition (CVPR). pp. 11372--11381 (2020).
  \doi{10.1109/CVPR42600.2020.01139}

\bibitem{yen2020inerf}
Yen-Chen, L., Florence, P., Barron, J.T., Rodriguez, A., Isola, P., Lin, T.Y.:
  {iNeRF}: Inverting neural radiance fields for pose estimation. In: IEEE/RSJ
  International Conference on Intelligent Robots and Systems ({IROS}) (2021)

\end{thebibliography}
\end{document}


\pagestyle{headings}
\mainmatter
\def\ECCVSubNumber{5158}  

\title{Camera Pose Auto-Encoders for Improving Pose Regression: Supplementary Materials} 

\titlerunning{Camera Pose Auto-Encoders: Supplementary Materials}
%
\author{Yoli Shavit \and
Yosi Keller}
\authorrunning{Y. Shavit and Y. Keller}
%
\institute{Bar-Ilan University, Ramat Gan, Israel\\
\email{\{yolisha, yosi.keller\}@gmail.com}}

\maketitle

\section{Appendix}
\subsection{Memory Requirements of APRs and RPRs}
A key motivation for PAEs is to reduce the memory burden associated with RPRs, which require train images or their encoding to be available at inference time. Table \ref{table:storage} shows the memory requirements for RPRs, single and multi-scene APRs with and without PAEs.
\begin{table*}[h!]
\caption{Order of magnitude of the storage required for different APRs and RPRs (considering 10 scenes). For RPRs, we assume that a single encoding weighs 5Kb and each scene contains 2000 images.}
\centering
\begin{tabular}
{L{4em} C{4em} C{4em}}\hline
\textbf{Method} &  \textbf{Storage} \\\hline
\multicolumn{1}{l}{RPR (\cite{balntas2018relocnet}) } & Gb \\
\multicolumn{1}{l}{Single Scene APR \cite{kendall2015posenet}} & Mb  \\\multicolumn{1}{l}{Multi Scene APR } & Mb \\
\multicolumn{1}{l}{Multi Scene APR + PAE} & Mb
\\\hline
\end{tabular}
\label{table:storage}
\end{table*}
\subsection{Data Augmentation and Training}
When training our camera pose auto-encoder and during test-time optimization, we follow the same test-time data pre-processing used by \cite{kendall2015posenet}. Specifically, images are first resized, where the smaller edge is resized to $256$ pixels, and then a $224\times 224$ center crop is taken. When training teacher APRs, we follow a similar procedure but additionally apply random jitter to the brightness, contrast, and saturation and take a random crop (rather than the center one). To train our decoder, we used $64x64$ crops (rescaling is done to maintain the original ratio between scaling and resizing). In order to support easy reproduction of the results reported by other researchers, we provide training and evaluation code, pretrained models, dedicated configuration files, and examples to perform each experiment.
\subsection{Additional Ablations}
\subsubsection{Ablation of Fourier Features}
We carry additional ablation on the number of periodic encoding functions $L$ (see Section 3.2 in the main text) for our Fourier Features. Table \ref{tb:ablation-fourier} shows the results for a PAE with a 4-layers MLP trained without Fourier Features and for a PAE with a 4-layers MLP trained with Fourier Features with $L=3$ and $L=6$. The latter configuration, which yields the lowest position error, is selected for our PAE architecture.
\begin{table*}[h!]
\caption{Ablations of Fourier Features for the PAE architecture . We compare the
median position and orientation errors when using a 4-layer MLP without Fourier Features and when applying Fourier Features with $L$ the number of levels set to 3 and 6 (selected architecture). The performance is
reported for the KingsCollege scene (CambridgeLandmarks dataset). The
Teacher is a PoseNet APR with a MobileNet architecture.}
\label{tb:ablation-fourier}\centering{\small
\begin{tabular}{lll}
\hline
\textbf{Auto Encoder Architecture} & \textbf{Position [m]} & \textbf{%
Orientation [deg]} \\ \hline
4-Layers MLP (No Fourier Features)& \multicolumn{1}{c}{1.26} & \multicolumn{1}{c}{3.54} \\
4-Layers MLP + Fourier Features, $L=3$ & \multicolumn{1}{c}{{1.36}} &
\multicolumn{1}{c}{3.27} \\
4-Layers MLP + Fourier Features, $L=6$ & \multicolumn{1}{c}{{1.15}} &
\multicolumn{1}{c}{3.58} \\\hline
\end{tabular}%
}
\end{table*}

\subsubsection{Architecture Choices}
\begin{itemize}
    \item Dimension of $\mathbf{\hat{z}_{x}}$ and $\mathbf{\hat{z}_{q}}$:
    the dimension of the PAE's latent vectors should match the dimension of the latent output of the APR teacher. For the APRs used in our paper, the dimension is the same for both vectors.
    \item Separate branches for position and orientation estimation: In our work, we use both single- and multi- scene APRs with separate branches for position and orientation. Nevertheless, PAEs can be applied to any APR architecture.
    \item Image size (image decoding): The choice of $64\times64$ pixels as the size of the reconstructed image is set to maintain a short runtime. We note that similar results (in terms of position error and image quality) were achieved with a higher image resolution (256 pixels).
\end{itemize}
\subsection{Test-time Position Refinement: Additional Results}
\begin{table*}[h!]
\caption{Median position error in meters when sampling a random guess around the ground-truth pose and when refining the initial guess with our test-time optimization. We report the results for the CambridgeLandmarks dataset.}
\label{tb:test-time-verfication}\setlength{\tabcolsep}{4pt} \centering
\begin{tabular}{cccccc}
\hline
\textbf{Method} & \textbf{K. College} & \textbf{Old Hospital} & \textbf{Shop
Facade} & \textbf{St. Mary} &  \\ \hline
\multicolumn{1}{l}{Initial Guess} & 1.47 & 1.45 & 1.53  & 1.8  \\ 
\multicolumn{1}{l}{Refined Guess} & 0.59 & 0.57 & 0.56 & 0.4 &  \\ \hline
\end{tabular}%
\end{table*}

\begin{table*}[h!]
\caption{Median position error of single-scene APRs, with and without our test-time position refinement. Performance is reported for
the KingsCollege scene (CambridgeLandmarks dataset).}
\label{table:test-time-single}\centering{\small
\begin{tabular}{lcc}
\hline
\textbf{APR Architecture} & \textbf{Without Position} &  \textbf{With Position} \\ & \textbf{ Refinement} [m] &   \textbf{ Refinement} [m] \\  \hline
PoseNet+MobileNet & 1.24 &  0.91 \\
PoseNet+ResNet50 & 1.56 &  1.31 \\
PoseNet+EfficientNet & 0.88 & 0.81  \\ \hline
\end{tabular}%
}\vspace{-0em}
\end{table*}

\subsubsection{Test-time Position Refinement with a Random Pose Guess}
We perform an additional verification of our test-time optimization, where instead of using an APR to estimate the pose of the query and its latent representation, we take a random guess around the ground truth pose and encode it (i.e., perform pose estimation \textit{without} images). Table \ref{tb:test-time-verfication} reports the results for the CambridgeLandmarks dataset, showing the accuracy of the position of the initial guess and the refined estimate, obtained with our test-time optimization. Our method can significantly reduce the error of the initial guess.
\subsubsection{Test-time Position Refinement with Single-scene APRs and PAEs}
We apply our test time position refinement to single scene APRs and their respective student PAEs, trained on the KingsCollege scene from the CambridgeLandmarks dataset.  Table \ref{table:test-time-single} shows the position error achieved by applying each single scene APR with and without our PAE-based position refinement. Our test-time optimization yields a consistent improvement, regardless of the APR architecture used. \subsection{Test-time Orientation Estimation with Affine Combination}
Our test-time refinement focuses on position estimation through affine combination of train positions, fetched based on PAE encoding. We further evaluate this procedure to refine the orientation estimation. Table \ref{tb:cambridge_orientation} shows the results for MS-Transformer with and without applying our affine combination to orientation estimation for the CambridgeLandmarks dataset. The affine combination leads to degradation, suggesting that additional research is needed to extend the PAE-based test time refinement to improve orientation estimation. Natural extensions are estimating the weights for position and orientation separately as well as combining the quaternions through quaternion averaging algorithms such as \cite{markley2007averaging} (rather than directly applying a weighted average as done in our proposed procedure).
\begin{table*}[h!]
\caption{Median orientation error in degrees for the CambridgeLandmarks dataset, obtained with MS-Transformer\cite{ShavitFerensIccv21} with and without the proposed test-time affine combination. }
\label{tb:cambridge_orientation}\centering\setlength{\tabcolsep}{4pt} {\small
\begin{tabular}{cccccc}
\hline
\textbf{Method} & \textbf{K. College} & \textbf{Old Hospital} & \textbf{Shop
Facade} & \textbf{St. Mary} &  \\ \hline
\multicolumn{1}{l}{MS-Transformer} & {1.47} & {2.39} & {%
3.07} & {3.99} &  \\
\multicolumn{1}{l}{MS-Transformer with} & 2.83 & 4.04 & 3.44 & 7.96  \\
Affine Combination & & & & \\
\hline
\end{tabular}%
}
\end{table*}
\subsection{Comparison of Camera Localization Methods}
Our work focuses on encoding camera poses and demonstrating their usages for absolute pose regression. However, absolute pose regression is one family of methods out of several clusters of techniques for camera localization, namely: structure-based methods, image retrieval, and relative pose regression (see our Related Work section). In order to support a more complete comparison, Table \ref{tb:comparison} shows the results for representative methods for the CambridgeLandmarks and 7Scenes datasets. Structure-based method achieve the best localization accuracy. However, they require the intrinsics of the query camera, which might not be accurate or available.

\begin{table*}[h!]
\caption{Comparison of localization methods when applied to the Cambridge Landmarks and 7Scenes datasets. We show results for representative methods from each localization family: structure-based (STR), image retrieval (IR), relative pose regression (RPR) and image based absolute pose regression (APR).
We report the average of median position/orientation errors across scenes in meters/degrees for each
method. }
\label{tb:comparison}\centering{\small
\begin{tabular}{llcc}
\hline
&  Method & \textbf{CambridgeLand.} & \textbf{7Scenes} \\ \hline\hline
\multirow{2}{*}{\rotatebox[origin=c]{90}{STR}} & DSAC~\cite{DSAC} & 0.15/0.4 & 0.03/1.4\\
& DSAC*~\cite{9394752} & 0.15/0.4 & --/--\\
\hline
\multirow{2}{*}{{\rotatebox[origin=c]{90}{IR}}} & VLAD~\cite{denseVLAD} &
2.56/7.1 & 0.26/12.5 \\
& VLAD+Inter~\cite{sattler2019understanding} & 1.67/4.9 &  0.24/11.7 \\ \hline
\multirow{5}{*}{{\rotatebox[origin=c]{90}{RPR}}} & EssNet~\cite{essnet} & 1.08/3.4 & 0.22/8.0\\
& VLocNet~\cite{valada2018deep} &
0.78/2.8 & 0.05/3.8\\
& GL-Net~\cite{glnet} & 1.22/2.4 & 0.19/6.3\\
& NC-EssNet~\cite{essnet} & 0.85/2.8 & 0.21/7.5\\
& RelocGNN\cite{9394752} & 0.91/2.3 & 0.91/2.3\\ \hline
\multirow{12}{*}{{\rotatebox[origin=c]{90}{APR}}} & PoseNet \cite%
{kendall2015posenet} & 2.09/6.84 & 0.44/10.4\\
& BayesianPN \cite{kendall2016modelling} & 1.92/6.28 & 0.47/9.81 \\
& LSTM-PN \cite{walch2017image} & 1.30/5.52 & 0.31/9.86
\\
& SVS-Pose \cite{naseer2017deep} & 1.33/5.17 & $--$
\\
& GPoseNet \cite{cai2019hybrid} & 2.08/4.59 & 0.31/9.95
\\
& PoseNetLearn \cite{kendall2017geometric} & 1.43/2.85 & 0.24/7.87 \\
& GeoPoseNet \cite{kendall2017geometric} & 1.63/2.86 & 0.23/8.12 \\
& MapNet \cite{brahmbhatt2018geometry} & 1.63/3.64 & 0.21/7.78
\\
& IRPNet \cite{shavitferensirpnet} & 1.42/3.45 & 0.23/8.49
\\
& AttLoc\cite{wang2020atloc} & $--$ & 0.20/7.56
\\
& MS-TransFormer\cite{ShavitFerensIccv21} & 1.28/2.73 & 0.18/7.28\\
& MS-Transformer + Position Refinement & 0.96/2.73 & 0.15/7.28 \\ \hline
\end{tabular}
}
\end{table*}

\bibliographystyle{splncs04}
\bibliography{localization}